\relax

\documentclass[twoside,11pt]{article}

\usepackage{jmlr2e}

\usepackage{blindtext}
\usepackage{graphicx}

\usepackage{times}
\usepackage{helvet}
\usepackage{courier}
\usepackage{url}
\usepackage{graphicx}
\usepackage{xcolor} 
\usepackage{algorithm}
\usepackage[noend]{algpseudocode}
\usepackage{graphicx}
\usepackage{subcaption}
\usepackage[utf8]{inputenc} 
\usepackage{booktabs}
\usepackage{amsfonts}
\usepackage{amsmath}
\usepackage{amssymb}
\usepackage{graphicx}
\usepackage{bm}
\usepackage{dsfont}
\usepackage{mathtools}

\title{Sum-Product Networks for Hybrid Domains}
\author{sum-product networks, non-parametric density estimation, non-parameteric independency test, hybrid domains, mixed graphical models}

\author{\name Alejandro Molina {*}
    \email first.last@tu-dortmund.de \\
    \addr CS Department \\
    TU Dortmund, Germany
    \AND
    \name Antonio Vergari {*}
    \email first.last@uniba.it \\
    \addr CS Department \\
    University of Bari, Italy
    \AND
    \name Nicola Di Mauro 
    \email first.last@uniba.it \\
    \addr CS Department \\
    University of Bari, Italy
    \AND
    \name Sriraam Natarajan 
    \email natarasr@indiana.edu \\
    \addr CS Department \\
    Indiana University, USA
    \AND
    \name Floriana Esposito 
    \email first.last@uniba.it \\
    \addr CS Department \\
    University of Bari, Italy
    \AND
    \name Kristian Kersting
    \email last@cs.tu-darmstadt.de \\
    \addr CS Dept.~and Centre for CogSci \\
    TU Darmstadt, Germany
    \\
    \\ * These authors contributed equally to this work
}

\newcommand{\argmax}{\operatornamewithlimits{argmax}}

\newcommand{\X}{\ensuremath{\mathbf{X}}}
\newcommand{\x}{\ensuremath{\mathbf{x}}}
\newcommand{\Y}{\ensuremath{\mathbf{Y}}}
\newcommand{\y}{\ensuremath{\mathbf{y}}}

\definecolor{eqGray}{HTML}{777777}
\definecolor{eqBlack}{HTML}{000000}

\definecolor{lacamdarklilac5} {RGB} {51, 10, 102}
\definecolor{lacamgold5} {RGB} {255, 87, 0}
\definecolor{violet} {RGB} {119, 111, 178}
\definecolor{petroil2} {RGB} {36, 165, 175}
\definecolor{petroil4} {RGB} {30, 132, 149}
\definecolor{petroil6} {RGB} {23, 101, 115}
\definecolor{gold2} {RGB} {255, 130, 0}
\definecolor{gold4} {RGB} {250, 100, 0}
\definecolor{gold6} {RGB} {245, 90, 0}

\begin{document}

\editor{}
\maketitle

\begin{abstract}
  While all kinds of mixed data---from personal data, over panel and scientific data, to public and commercial data---are collected and stored, building probabilistic graphical models for these hybrid domains becomes more difficult. Users spend significant amounts of time in identifying the parametric form of the random variables (Gaussian, Poisson, Logit, etc.) involved and learning the mixed models. 
  To make this difficult task easier, we propose the first trainable probabilistic deep architecture for hybrid domains that features tractable queries. It is based on Sum-Product Networks (SPNs) with piecewise polynomial leave distributions together with novel nonparametric decomposition and conditioning steps using the Hirschfeld-Gebelein-R\'{e}nyi  Maximum Correlation Coefficient. This relieves the user from deciding a-priori the parametric form of the random variables but is still expressive enough to effectively approximate any continuous distribution and permits efficient learning and inference. Our empirical evidence shows that the architecture, called Mixed SPNs, can indeed capture complex distributions across a wide range of hybrid domains.
\end{abstract}

\section{Introduction}

Machine learning has achieved considerable successes in recent years and an ever-growing number of disciplines rely on it. Data is now ubiquitous, and there is great value from understanding the data, building probabilistic models and making predictions with them. However, in most cases, this success crucially relies on the data scientists to posit the right parametric form of the probabilistic model underlying the data, to select a good algorithm to fit to their data, and finally to perform inference on it.

These can be quite challenging even for experts and often go beyond non-experts' capabilities, specifically in hybrid domains, consisting of mixed---continuous, discrete and/or categorical---statistical types. Building a probabilistic model that is both expressive enough to capture complex dependencies among random variables of different types as well as allows for effective learning and efficient inference is still an open problem.

More precisely, most existing graphical models for hybrid domains---also called \emph{mixed models}---are limited to particular combinations of variables of parametric forms such as the Gaussian--Ising mixed model~\cite{Lauritzen1989}, where there are Gaussian and multinomial random variables, and the continuous variables are conditioned on all configurations of the discrete variables. Unfortunately,  inference in this Gaussian-Ising mixed graphical model scales exponentially with the number of discrete variables, and only recently, 3-way dependencies have been realized~\cite{Cheng2014}. Therefore it is not surprising that hybrid Bayesian networks (HBNs) have restricted their attention to simpler parametric forms for the conditional distributions such as conditional linear Gaussian models~\cite{Heckerman1995}.

While extensions based on copulas aim to provide more flexibility~\cite{Elidan2010}, selecting the best parametric copula distribution for each application requires a significant engineering effort. 
Probably the most recent approach are Manichean graphical models~\cite{Yang2014}, and we refer to this paper for an excellent recent overview on mixed graphical models. Manichean models---after the philosophy that loosely places elements into one of two types---specify that each of the conditional distributions is a member of a possibly different univariate exponential family. Although indeed more flexible than Gaussian-Ising mixed models, Manichean models are still demanding, in particular when it comes to inference. Alternatively, one may make a piecewise approximation to continuous distributions~\cite{Shenoy2011}. In their simplest form, piecewise constant functions are often adopted in the form of histograms or staircase functions, and more expressive approximations comprise mixtures of truncated polynomials~\cite{Langseth2012} and exponentials~\cite{Moral2001}. This has resulted 
in a number of novel inference approaches for hybrid domains \cite{Sanner2012,Belle2015a,Belle2015b,morettinPS17}.

Although expressive, in particular learning these non-parametric models has not been considered or does not scale. To overcome the difficultness of mixed probabilistic graphical modeling and inspired by the successes of deep models, we introduce Mixed Sum-Product Networks (MSPNs). They are a general class of mixed probabilistic models that, by combining Sum-Product Networks~\cite{Poon2011} and piecewise polynomials, allow for a large range of exact and tractable inference without making distributional assumptions.
Learning MSPNs from data, however, requires novel decomposition and conditioning steps for Sum-Product Networks (SPNs) tailored towards nonparametric distributions. Providing them based on the R\'{e}nyi Maximum Correlation Coefficient~\cite{LopezPaz2013}---the first application of it to learning sum-product networks---via a series of variable transformations is our main technical contribution.
This then naturally results in the first automated tool for learning multivariate distributions over hybrid domains without requiring users to decide the parametric form of random variables or their dependencies, yet enabling them to answer complex probabilistic queries efficiently on tasks previously unfeasible by classical mixed models.
We proceed as follows. We start off by reviewing SPNs. Afterwards, we introduce MSPNs and show how to learn tree-structured MSPNs from data using the R\'{e}nyi Maximum Correlation Coefficient. Before concluding, we present our experimental evaluation. 

\section{Sum-Product Networks (SPNs)}
Recent years have seen a significant interest in tractable probabilistic representations such as Arithmetic Circuits (ACs), see \cite{choiD17} for a discussion. In particular, SPNs, an instance of ACs, are deep probabilistic models that can represent high-treewidth models \cite{Zhao2015} and facilitate \emph{exact} inference for a range of queries in time \emph{polynomial} in the network size \cite{Poon2011,Bekker2015}.

\textbf{Definition of SPNs:} 
Formally, an SPN  is a rooted directed acyclic graph, comprising \emph{sum}, \emph{product} or \emph{leaf} nodes. The scope of an SPN is the set of random variables appearing in the network. An SPN can be defined recursively as follows:
(1) a tractable univariate distribution is an SPN;
(2) a product of SPNs defined over different scopes is an SPN; and
(3), a convex combination of SPNs over the same scope is an SPN. 
Thus, a product node in an SPN represents a factorization over independent distributions defined over different random variables, while a sum node stands for a mixture of distributions defined over the same variables. From this definition, it follows that the joint distribution modeled by such an SPN is a valid probability distribution, i.e., each complete and partial evidence inference query produces a consistent probability value~\cite{Poon2011,Peharz2015a}. This also implies that we can construct multivariate distributions from simpler univariate ones. Furthermore, any node in the network could be replaced by any
tractable multivariate distribution over the same scope, obtaining still a valid SPN.

\textbf{Tractable Inference in SPNs:}
To answer probabilistic queries in an SPN, we evaluate the nodes starting at the leaves. Given some evidence, the probability output of querying leaf distributions is propagated bottom up. For product nodes, the values of the children nodes are multiplied and propagated to their parents. For sum nodes, instead, we sum the weighted values of the children nodes. The value at the root indicates the probability of the asked query. To compute marginals, i.e., the probability of partial configurations, we set the probability at the leaves for those variables to $1$ and then proceed as before. Conditional probabilities can then be computed as the ratio of partial configurations. To compute MPE states, we replace sum by max nodes and then evaluate the graph first with a bottom up pass, but instead of weighted sums we pass along the weighted maximum value. Finally, in a top down pass, we select the paths that lead to the maximum value, finding approximate MPE states \cite{Poon2011}. All these operations traverse the tree at most twice and therefore can be achieved in linear time w.r.t. the size of the SPN.

\begin{figure*}[t] 
\centering
 \includegraphics[width=0.48\linewidth]{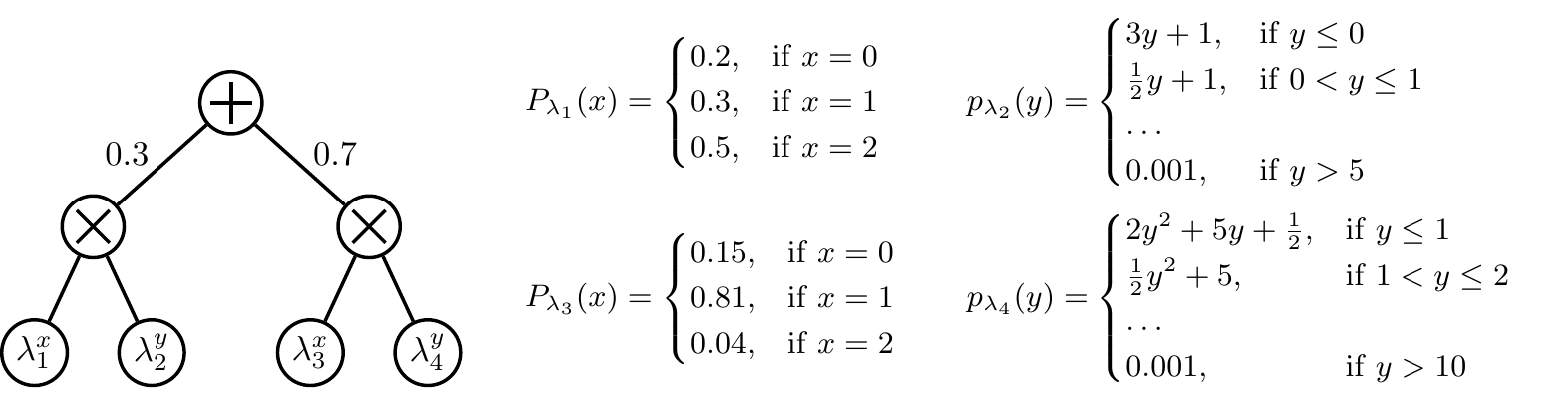}
 \includegraphics[width=0.24\linewidth]{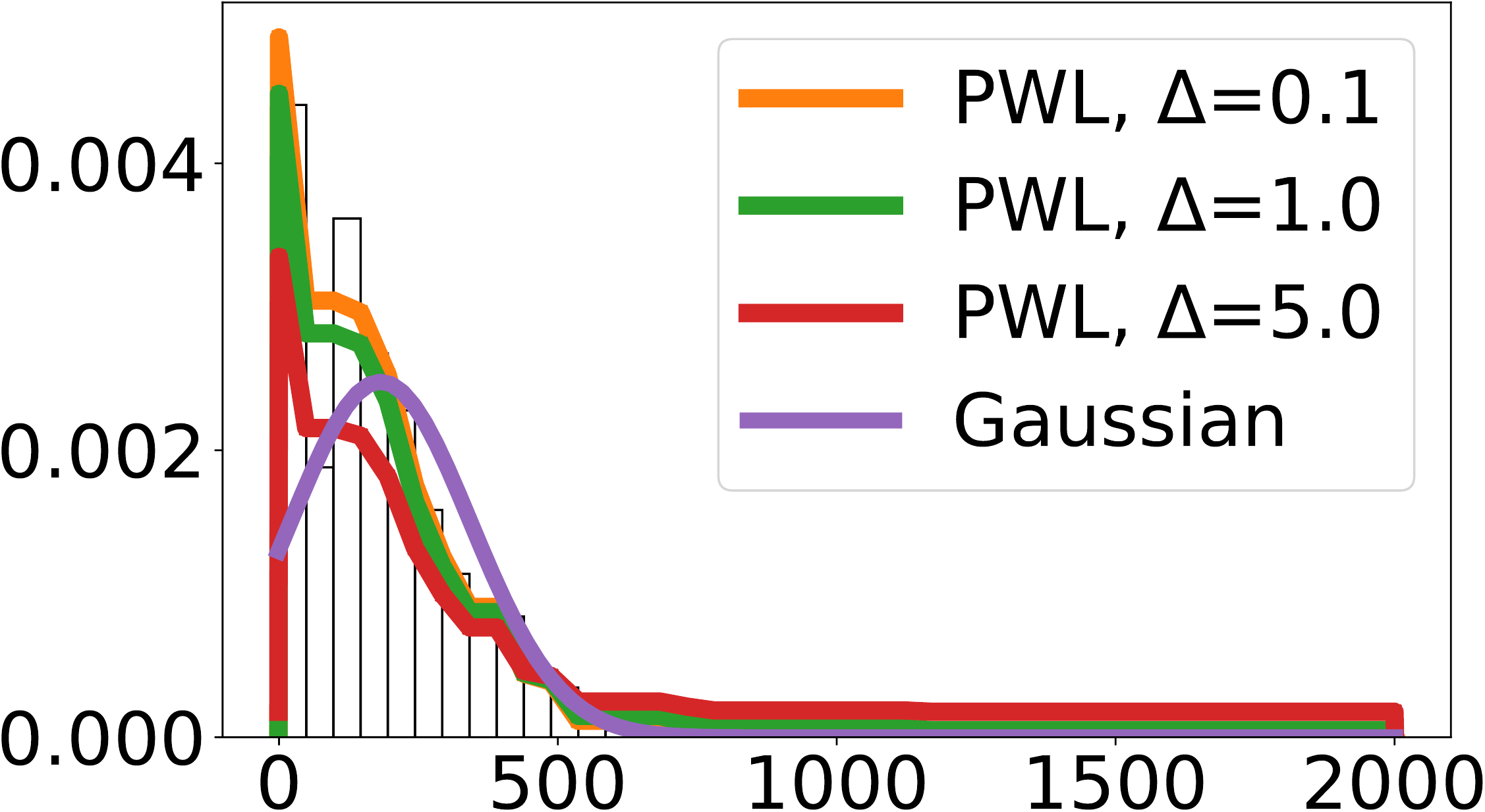}
  \includegraphics[width=0.24\linewidth]{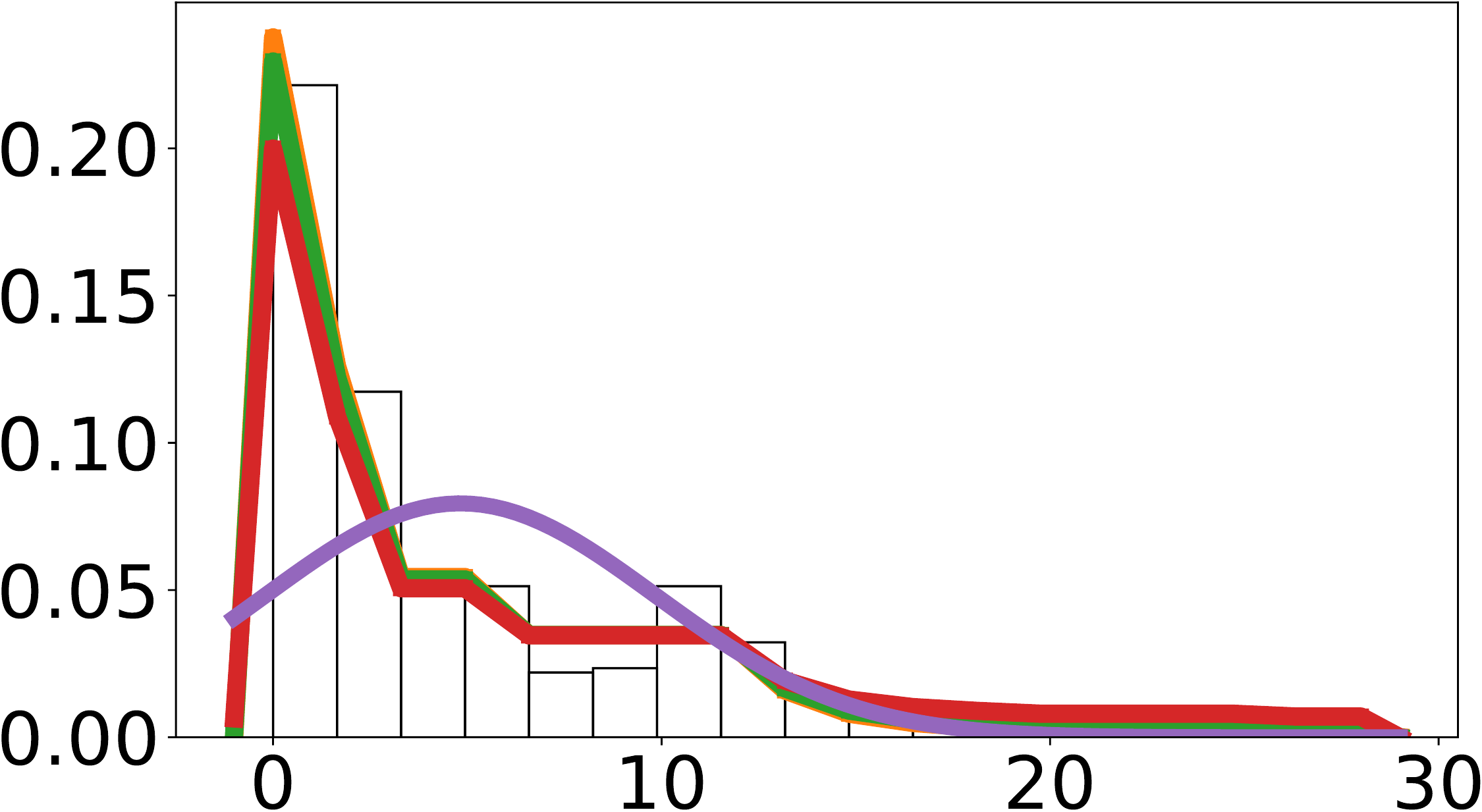}
 \caption{
   Mixed Sum-Product Networks (MSPNs). From left to right: {\bf (left)} An MSPN representing a mixture over random variables $x$ (discrete)
   and $y$ (continuous).
   {\bf (middle)} In each each leaf $\lambda_{i}$ an MSPN approximates the univariate distribution as a      piecewise polynomial. In our experiments we employ piecewise linear models {\bf (right)}. Shown are fitted distributions on the UCI Australian dataset, see experiment section. Shown are the empirical distributions (histograms), piecewise linear approximations using isotonic regression (PWL) with different smoothing values ($\Delta$), and superimposed maximum likelihood Gaussians. This shows that piecewise polynomials are expressive enough to effectively approximate continuous distributions, often     better than the classical Gaussian assumption. Within a MSPN, they permit efficient learning and inference. (Best viewed in color)
 \label{fig:Inference}\label{fig:piecewise-approx}} 
\end{figure*}

\textbf{Learning SPNs:}
While it is possible to craft a valid SPN structure by hand, doing so would require domain knowledge and weight learning afterwards~\cite{Poon2011}. Here, we focus on a top-down approach~\cite{Lowd2008,Gens2013} that directly learns both the structure and weights of (tree) SPNs at once. It uses three steps: (1) base case, (2) decomposition and (3) conditioning. In the base case, if only one variable remains, the algorithm learns a univariate distribution and terminates. In the decomposition step, it tries to partition the variables into independent components $V_j \subset \mathbf{V}$ such that $P(\mathbf{V}) = \prod\nolimits_{j}P\left(V_j\right)$ and recurses on each component, inducing a product node. If both the base case and the decomposition step are not applicable, then training samples are partitioned into clusters (conditioning), inducing a sum node, and the algorithm recurses on each cluster. 

This scheme for learning tree SPNs has been instantiated for several well-known  distributions with parametric forms. Conditioning for Gaussians can be realized using hard clustering with EM or K-means~\cite{Gens2013,Rooshenas2014-short}.
For Poissons, mixtures of Poisson Dependency Networks have been proven successful~\cite{Molina2017}. For the decomposition step, one typically employs pairwise independence tests with some associated independence score $\rho$. For categorical variables, \citeauthor{Gens2013}~(\citeyear{Gens2013}) proposed to use the G-test, and \citeauthor{Rooshenas2014-short}~(\citeyear{Rooshenas2014-short}) a pairwise mutual information test. For variables of the generalized linear model family, Molina {\it et al.}~(\citeyear{Molina2017}) proposed the use of parameter instability tests based on generalized M-fluctuation processes. Then, one creates an undirected graph where there is an edge between  random variables $V_i$ and $V_j$ if the value $\rho(V_{i}, V_{j})$ passes a threshold of significance $\alpha$. That is, the decomposition step equals to partitioning the graph into its connected components. It is rejected if there is only a single connected component.

\section{Mixed Sum-Product Networks (MSPNs)}\label{MSPN}
Unfortunately, all previous decomposition and conditioning approaches for SPNs are only suitable for multivariate distributions of known parametric form: categorical, binomial, Gaussian and Poisson distributions~\cite{Poon2011,Vergari2015,Molina2017}. To model hybrid domains without making parametric assumptions, one has to introduce new conditioning and decomposition approaches tailored towards mixed models.

\textbf{R{\'e}nyi Decomposition:}
We approach the problem of seeking independent subsets of random variables of mixed but unknown types as a dependency discovery problem. Alfred R{\'e}nyi~(\citeyear{Renyi1959}) argued that a measure of maximum dependence $\rho^{*} : V_{i} \times V_{j} \rightarrow [0, 1]$ between random variables $V_{i}$ and $V_{j}$ should satisfy several fundamental properties, such as symmetry, transformation invariance, and it should also hold that $\rho^{*}(V_{i}, V_{j})=0$ iff $V_{i}$ and $V_{j}$ are statistically independent. He also showed the Hirschfeld-Gebelein-Renyi (HGR) Maximum Correlation Coefficient due to \citeauthor{Gebelein1941}~(\citeyear{Gebelein1941}) to satisfy all these properties. Recently, \citeauthor{LopezPaz2013}~(\citeyear{LopezPaz2013}) provided a practical estimator for the HGR $\rho^{*}$, the randomized dependency coefficient (RDC). The RDC is appealing for hybrid domains because it can be applied to both multivariate, continuous and discrete random variables. Also, its $\mathcal{O}(M\log{M})$ running time, with $M$ being the number of instances, makes it one of the fastest non-linear dependency measures.

The general idea behind the RDC is to look for linear correlations between the representations of two random samples that have undergone a series of non-linear transformations. The two samples are deemed statistically independent iff the transformed samples are linearly uncorrelated. This is the same reasoning behind the adoption of higher space projections for the kernel-trick in classification and the stacking of representations in deep architectures.

Specifically, consider two random samples $\mathcal{D}_{V_{i}}=\{v^{m}_{i}|v^{m}_{i}\sim V_{i}\}_{m=1}^{M}$ and $\mathcal{D}_{V_{j}}=\{v^{m}_{j}|v^{m}_{j}\sim V_{j}\}_{m=1}^{M}$ drawn from variables  $V_{i}$ and $V_{j}$, we decide that $V_{i}$ and $V_{j}$ are independent iff  $\rho(\mathcal{D}_{V_{i}}, \mathcal{D}_{V_{j}})=0$, where $\rho$ is the RDC.

Instead of operating directly on $\mathcal{D}_{V_{i}}$ and $\mathcal{D}_{V_{j}}$, and in order to achieve invariance against scaling and shifting data transformations, we first compute their \emph{empirical copula transformations}~\cite{Poczos2012}, $\mathcal{C}_{V_{i}}$ and respectively $\mathcal{C}_{V_{j}}$, in the following way: 
\begin{equation}
  \mathcal{C}_{V_{i}}=\bigg\{ \frac{1}{M}\sum\nolimits_{r=1}^{M}\mathds{1}\{v_{i}^r \leq v_{i}^{m}\}\bigg|v^{m}_{i}\in\mathcal{D}_{V_{i}}\bigg\}_{m=1}^{M}
  \label{eq:copula}
\end{equation}
Then, we apply a random linear projection on the obtained samples to a $k$-dimensional space, finally passing them through a non-linear function $\sigma$.
We compute: 
\begin{equation}
  \phi(\mathcal{C}_{V_{i}})=\sigma(\mathbf{w}\cdot\mathcal{C}_{V_{i}}^{T}+b), (\mathbf{w}, b)\sim\mathcal{N}(\mathbf{0}_{k}, s\mathbf{I}_{k\times k})
  \label{eq:nonlinear}
\end{equation}
for the first sample, an equivalent transformation yields $\phi(\mathcal{C}_{V_{j}})$.

Note that $\mathbf{w}\in\mathbb{R}^{k\times 1}$, $b\in\mathbb{R}$ and that random sampling  $\mathbf{w}$ from a zero-mean $k$-dimensional Gaussian is analogous to the use of a Gaussian kernel~\cite{Rahimi2009}.
We choose $k=20$, $\sigma$ to be sine function and $s=\frac{1}{6}$ as both have proven to be reasonable empirical heuristics, see~\cite{LopezPaz2013}. Lastly, we compute the canonical correlations ($\text{CCA}$) $\rho^{2}$ for $\phi(\mathcal{C}_{V_{i}})$ and $\phi(\mathcal{C}_{V_{j}})$ as the solutions for the following eigenproblem:
\begin{equation}
  \begin{pmatrix}
0 & \Sigma^{-1}_{ii}\Sigma_{ij} \\ 
\Sigma^{-1}_{jj}\Sigma_{ji} & 0
\end{pmatrix}
\begin{pmatrix}
\beta \\ 
\gamma
\end{pmatrix}=\rho^2
\begin{pmatrix}
\beta \\ 
\gamma
\end{pmatrix},
\label{eq:cancor}
\end{equation}
where the covariance block matrices involved are: 
$$\Sigma_{ij}=\text{cov}(\phi(\mathcal{C}_{V_{i}}),\phi(\mathcal{C}_{V_{j}})), \Sigma_{ji}=\text{cov}(\phi(\mathcal{C}_{V_{j}}),\phi(\mathcal{C}_{V_{i}})),$$
$$\Sigma_{ii}=\text{cov}(\phi(\mathcal{C}_{V_{i}}),\phi(\mathcal{C}_{V_{i}})), \Sigma_{jj}=\text{cov}(\phi(\mathcal{C}_{V_{j}}),\phi(\mathcal{C}_{V_{j}})).$$
In the end, the actual value for the RDC coefficient is the largest canonical correlation coefficient:
\begin{equation}
  \text{RDC}(V_{i},V_{j}) =\sup\nolimits_{\beta, \gamma}\rho(\beta^{T}\phi(\mathcal{C}_{V_{i}}),\gamma^{T}\phi(\mathcal{C}_{V_{j}})).
\end{equation}

This RDC pipeline goes through a series of data transformations, which constitutes the basis of our decomposition procedure, cf.~Alg.~\ref{algo:splitFeatures}. We note that all the transformations presented so far are easily generalizable to the multivariate case~\cite{LopezPaz2013}. We are applying these multivariate versions both when performing conditioning on multivariate samples (see below) and when we deal with decomposing categorical random variables. Since Eq.~\ref{eq:copula} is not well defined for categorical data, to treat them in the same way as continuous and discrete data, we proceed as follows. First we perform a one hot encoding transformation for each categorical random variables $V_{c}$, obtaining a multivariate binary random variable $\mathbf{B}_{V_{c}}$. 
Then, we apply Eq.~\ref{eq:copula} to each column $\mathbf{B}_{V_{c}}$ independently, obtaining the matrix $\mathcal{C}_{\mathbf{B}_{V_{c}}}$.
This way we are preserving all the modalities of $V_{c}$. Finally, we apply the generalized version of Eq.~\ref{eq:nonlinear} and Eq.~\ref{eq:cancor} to the multivariate case. 

Note that, while we are looking for the RDC to be zero in case of independent random variables, it is extremely unlikely for this to happen on real data samples. In practice, the thresholding approach on the adjacency graph induced by dependencies (see the previous section) takes care of this for the decomposition step.

\begin{algorithm}[t]
  \caption{\textsf{splitFeaturesRDC} ($\mathcal{D}$, $\alpha$)}
  \label{algo:splitFeatures}
  \begin{algorithmic}[1]
    \State \textbf{Input:} samples $\mathcal{D}=\{\mathbf{v}^{m}=(v_{1}^{m},\dots,v_{N}^{m})|\mathbf{v}^{m}\sim \mathbf{V}\}_{m=1}^{M}$ over a set of
    random variables $\mathbf{V}=\{V_{1},\dots,V_{N}\}$; $\alpha$: threshold of significance
    \State \textbf{Output:} a feature partition $\{\mathcal{P}_{\mathcal{D}}\}$
    \For {\textbf{each} $V_{i}\in \mathbf{V}$}
    \State $\mathcal{C}_{V_{i}}\leftarrow\bigg\{ \frac{1}{M}\sum_{r=1}^{M}\mathds{1}\{v_{i}^{r} \leq v_{i}^{m}\}\bigg|v^{m}_{i}\in\mathcal{D}_{V_{i}}\bigg\}_{m=1}^{M}$
    \State $(\mathbf{w_i},b_i)\sim\mathcal{N}(\mathbf{0}_{k}, s\mathbf{I}_{k\times k})$
    \State $\phi(\mathcal{C}_{V_{i}})\leftarrow\sin(\mathbf{w_i}\cdot\mathcal{C}_{V_{i}}^{T}+b_i)$
    \EndFor
     \State $\mathcal{G}\leftarrow\mathsf{Graph(\{\})}$ 
    \For {\textbf{each} $V_{i}, V_{j}\in \mathbf{V}$}
        \State $c_{i, j}\leftarrow \mathsf{CCA}(\phi(\mathcal{C}_{V_{i}}),\phi(\mathcal{C}_{V_{j}}))$
        \If {$c_{i,j} > \alpha$}
            \State $\mathcal{G}\leftarrow\mathcal{G}\cup\{(i, j)\}$
        \EndIf
    \EndFor\\
  \Return $\mathsf{ConnectedComponents}(\mathcal{G})$
  \end{algorithmic}
\end{algorithm}

\textbf{R{\'e}nyi Conditioning:}
The task of clustering hybrid data samples depends on the choice of the metric space, which in turn, typically depends on the parametric assumptions made for each variable. Consider e.g. the popular choice of K-Means using the Euclidean metric. It makes a Gaussian assumption and therefore is not principled for categorical data.
To eliminate the reliance on knowing the type, we propose to cluster multivariate hybrid samples after they have been processed by the RDC pipeline. Not only does the series of non-linear transformations produce a feature space in which clusters may be more easily separable, but no distributional assumptions are required. More formally, given a set of samples $\mathcal{D}$ over RVs $\mathbf{V}$ we split it into a sample partitioning $\mathcal{P}_{\mathcal{D}}=\{\mathcal{D}_{c}\}_{c=1}^{C}$, $\bigcup_{c=1}^{C}\mathcal{D}_{C}=\mathcal{D}$, and $ \mathcal{D}_{q}\cap\mathcal{D}_{r}=\emptyset,\forall\mathcal{D}_{q},\mathcal{D}_{r}\in\mathcal{P}_{\mathcal{D}}$. The weights for the convex combination on the sum nodes are estimated as the proportions of the data belonging to each cluster, i.e., $w_c = \frac{|\mathcal{D}_c|}{|\mathcal{D}|}$.%
The procedure is sketched in Alg.~\ref{algo:clusterSamples}. First, we transform every feature $V_{i}$ in ${\mathcal{D}}$ using Eq~\ref{eq:nonlinear}:
 ${\mathcal{E}}=\{\phi(\mathcal{D}_{V_{n}})|\mathcal{D}_{V_{n}}\}_{n=1}^{N}.$
Then, all our features are projected into  a new $k$-dimensional non-linear space.  In this new space we can safely apply now K-Means to obtain $c$ clusters. In Alg.~\ref{algo:clusterSamples}, we set $c=2$ as this generally leads to deeper networks~\cite{Vergari2015}.

\begin{algorithm}[t]
  \caption{\textsf{clusterSamplesRDC} ($\mathcal{D}$)}
  \label{algo:clusterSamples}
  \begin{algorithmic}[1]
    \State \textbf{Input:} samples $\mathcal{D}=\{\mathbf{v}^{m}=(v_{1}^{m},\dots,v_{N}^{m})|\mathbf{v}^{m}\sim \mathbf{V}\}_{m=1}^{M}$ over a set of
    random variables $\mathbf{V}=\{V_{1},\dots,V_{N}\}$
    \State \textbf{Output:} a data partition $\{\mathcal{P}_{\mathcal{D}}\}$
    \State $\mathcal{C}_{V_{i}}\leftarrow\bigg\{ \frac{1}{M}\sum_{r=1}^{M}\mathds{1}\{v_{i}^{r} \leq v_{i}^{m}\}\bigg|v^{m}_{i}\in\mathcal{D}_{V_{i}}\bigg\}_{m=1}^{M}$
    \State $(\mathbf{w},b)\sim\mathcal{N}(\mathbf{0}_{s}, s\mathbf{I}_{k\times k})$
    \State $\phi(\mathcal{C}_{V_{i}})\leftarrow\sin(\mathbf{w}\cdot\mathcal{C}_{V_{i}}^{T}+b)$
    \State $\mathcal{E}\leftarrow \{\phi(\mathcal{C}_{V_{1}}),\dots,\phi(\mathcal{C}_{V_{N}})\}$\\
  \Return $\mathsf{KMeans}(\mathcal{E},2)$
  \end{algorithmic}
\end{algorithm}

\textbf{Nonparametric Univariate Leave Distributions:}
Finally, to be fully type agnostic, i.e., to realize MSPNs, we adopt piecewise polynomial approximations of the univariate leaf densities. The simplest and most straightforward approximation we consider are piecewise constant functions, i.e. histograms. 
More precisely, we adopt the scheme proposed in~\cite{Rozenholc2010} offering an adaptive binning, i.e. with irregular intervals, that is learned from data by optimizing a penalized likelihood function.
This allows MSPNs to model both multimodal and skewed univariate distributions without further assumptions. 
We apply Laplacian smoothing by a factor $\Delta$ to cope with empty bins and unseen values on the distribution domain.

\begin{algorithm}[t]
  \caption{\textsf{LearnMSPN} ($\mathcal{D}$, $\Delta$, $\eta$, $\alpha$)}
  \label{algo:learnaspn}
  \begin{algorithmic}[1]
    \State \textbf{Input:} samples $\mathcal{D}=\{\mathbf{v}^{m}=(v_{1}^{m},\dots,v_{N}^{m})|\mathbf{v}^{m}\sim \mathbf{V}\}_{m=1}^{M}$ over a set of
    random variables $\mathbf{V}=\{V_{1},\dots,V_{N}\}$; $\eta$: minimum
    number of instances to split; $\Delta$: histogram smoothing factor; $\alpha$: threshold of significance
    \State \textbf{Output:} an MSPN $S$ encoding a joint pdf over $\mathbf{V}$ learned from $\mathcal{D}$
    \If {$|\mathbf{V}| = 1$}
    \State $\{\mathcal{D}_{c}\}_{c=1}^C\leftarrow \mathsf{clusterSamplesRDC}(\mathcal{D})$
    \If {$C>1$}
    \State $S\leftarrow \sum_{i=1}^{C} \frac{|\mathcal{D}_c|}{|\mathcal{D}|}\mathsf{LearnMSPN}(\mathcal{D}_{c}, \Delta, \eta)$
    \Else
    \State $S\leftarrow \mathsf{LearnIsotonicLeaf}(\mathcal{D}, \Delta)$
    \EndIf
    \ElsIf {$|\mathcal{D}| < \eta$}
    \State $S \leftarrow \prod\limits_{n=1}^{|\mathbf{V}|}\mathsf{LearnMSPN}(\{v_{n}^{m}|v_n^{m}\sim V_n\}_{m=1}^{M}, \Delta, \eta)$
    \Else
    \State $\{\mathbf{V}_{c}\}_{c=1}^C\leftarrow \mathsf{splitFeaturesRDC}(\mathcal{D}, \alpha)$
    \If {$C > 1$}
    \State $\mathcal{D}_{c} \leftarrow \{\mathbf{v}_{c}^{m}|\mathbf{v}_c^{m}\sim \mathbf{V}_c\}_{m=1}^{M}$
    \State $S\leftarrow  \prod_{c=1}^{C}\mathsf{LearnMSPN}(\mathcal{D}_{c}, \Delta, \eta)$
    \Else
    \State $\{\mathcal{D}_{c}\}_{c=1}^C\leftarrow \mathsf{clusterSamplesRDC}(\mathcal{D})$
    \State $S\leftarrow \sum_{i=1}^{C} \frac{|\mathcal{D}_c|}{|\mathcal{D}|}\mathsf{LearnMSPN}(\mathcal{D}_{c}, \Delta, \eta)$
    \EndIf
    \EndIf
    \Return $S$
  \end{algorithmic}
\end{algorithm}

Indeed, by increasing the degree of leaf polynomial approximations, one can favor more expressive models. To balance between the complexity of learning resp.~inference and expressiveness, however, we adopt more complex models up to piecewise linear approximations.

We reframe the unsupervised task of estimating the density of univariate leaf distributions into a supervised one by fitting a nonparametric unimodal distribution function through isotonic regression~\cite{Frisen1986}, referred to as $\mathsf{LearnIsotonicLeaf}$. Once we have collected a set of pairs of points, e.g. from the previously estimated histogram, we employ them as labeled observations to fit a monotonically increasing (resp. decreasing) piecewise linear function up to (resp. down from) the estimated distribution mode. 
Note that the unimodality assumption for leaves is realistic, since we can accommodate \textsf{LearnMSPN}, cf.~Alg.~\ref{algo:learnaspn}, to grow a leaf only after no more clustering steps are possible, i.e. it is difficult if not impossible to separate two modalities in the observed data.

Now we have everything together to evaluate MSPNs empirically. Before doing so, we would like to stress that we here focused on a general setting. Instead of piecewise linear leaves, one can also employ existing hybrid densities as leave distributions such as HBNs, mixtures of truncated exponential families, or other nonparamteric density estimators such as Kernel Density Estimators (KDEs) and even denoising and variational autoencoders.

\section{Experimental Evaluation}
Our intention is to investigate the benefits of MSPNs compared to other mixed probabilistic models in terms of accuracy and flexibility of inference. Specifically, we investigate the following  questions:
\textbf{(Q1)} Is the MSPN distribution flexible for hybrid domains?
\textbf{(Q2)} How do MSPNs compare to existing mixed models?
\textbf{(Q3)} How do MSPNs compare to state-of-the-art parametric models in a single-type domain?
\textbf{(Q4)} Can MSPNs be applied across a wide range of distributions and inference tasks?
\textbf{(Q5)} Are there benefits of having tractable inference for hybrid domains, even via symbolic computation?

To this aim, we implemented MSPNs in Python calling R. All experiments ran on a Linux machine with 56 cores, 4 GPUs and 512GB RAM.

\begin{table}[!t]
  \centering
  \scriptsize
  \setlength{\tabcolsep}{3pt}  
  \resizebox{\columnwidth}{!}{
  \begin{tabular}{lrrrrr}
    & &\multicolumn{4}{c}{\bf MSPN}\\
    \cline{3-6}
    &&\multicolumn{2}{c}{\bf Gower}&\multicolumn{2}{c}{\bf RDC}\\
    \textbf{dataset}& $\mathsf{HBN}_{\mathsf{MMHC}}$&\textsf{hist}&\textsf{iso}&\textsf{hist}&\textsf{iso}\\
    \toprule
    \textsf{anneal-U} & -42.647 & -63.553 &  -38.836 & -60.314 &  \textbf{-38.312}\\ 
    \textsf{australian} & -38.423 & -18.513 & -30.379 & \textbf{-17.891} &  -31.021\\
    \textsf{auto} & -71.530 &-72.998 & \textbf{-69.405} & -73.378 &  -70.066\\ 
    \textsf{balance-scale} & -7.483 &-8.038 &  \textbf{-7.045} & -7.932 &  -7.302\\
    \textsf{breast} & -30.572 &-34.027 &  \textbf{-23.521} & -34.272 &  -24.035\\
    \textsf{breast-cancer} & \textbf{-9.193} &-15.373 &  -9.500 & -16.277 &  -9.990\\
    \textsf{cars} & \textbf{-28.596} & -30.467 &  -31.082 & -29.132 &  -30.516\\
    \textsf{cleave} & -26.296 & -26.132 &  -25.869 & -25.707 &  \textbf{-25.441}\\
    \textsf{crx} & -34.563 & \textbf{-22.422} &  -31.624 & -24.036 &  -31.727\\
    \textsf{diabetes} & -29.797 & \textbf{-15.286} &  -26.968 & -15.930 &  -27.242\\
    \textsf{german} & -34.356 & -40.828 &  -33.480 & -38.829 &  \textbf{-32.361}\\
    \textsf{german-org} & -29.051 & -43.611 & \textbf{-26.852} & -37.450 &  -27.294\\
    \textsf{heart} & -28.519 & -20.691 &  -26.994 & \textbf{-20.376} &  -25.906\\
    \textsf{iris} & \textbf{-1.670} &-3.616 &  -2.892 & -3.446 &  -2.843\\
  \bottomrule
  \bf wins over $\mathsf{HBN}_{\mathsf{MMHC}}$& - &\textbf{4/14} &\textbf{11/14} &\textbf{4/14} &\textbf{11/14}\\ \cline{3-6}
  \bf wins & \textbf{3/14}  & \multicolumn{4}{c}{\textbf{11/14}}

  \end{tabular}
  }
  \caption[datasets]{Average test set log likelihoods for UCI hybrid datasets (the higher the better). The best results are bold. MSPNs win in 11 out of 14 cases, even without information about the statistical types (RDC, iso). A Wilcoxon sign test shows that this is significant ($p=0.05$).\label{tab:uci-lls}}
  
\end{table}

\textbf{Hybrid UCI Benchmarks \textbf{(Q1, Q2)}:}
We considered the 14 preprocessed UCI benchmarks from the MLC++ library\footnote{\url{https://www.sgi.com/tech/mlc/download.html}} listed in Table~\ref{tab:uci-lls}. The domains span from survey data, to medical and biological domains, and they contain both continuous, discrete and categorical variables in different proportions.

As a baseline density estimator we considered HBNs whose conditional dependencies are modeled as conditional linear gaussians~\cite{Heckerman1995}. To learn their structure we explored both score-based and constrained-based approaches, finding the Max-Min Hill Climbing (MMHC) algorithm~\cite{Tsamardinos2006} to perform the best on the holdout data. For weight learning, we optimized the BDeu score. As an additional sanity check of our nonparametric RDC pipeline, we also trained MSPNs employing K-Medoids using the Gower distance (GowerMSPNs). The Gower distance~\cite{Gower1971} defines a metric over hybrid domains, at the cost of making distributional assumptions for each variable involved: take the average $d(i,j) = ({1}\slash{N})\sum_{n=1}^{N} d_{i,j}^{(n)}$ of distances $d_{i,j}^{(n)}$ per feature $n$. We assumed continuous variables to be Gaussian and discrete ones to be binomial.

The results are summarized in Table~\ref{tab:uci-lls}. MSPNs clearly outperform HBNs. Moreover, the performance of R{\'e}nyi conditioning is comparable to GowerMSPNs. This shows that using RDC is a sensible idea and frees the user from making parametric assumptions. Using histogram representations allows one to capture mixtures, which turns out to be beneficial for some datasets, but also results in a higher variance in performance across datasets, giving a benefit to isotonic regression. This answers {\bf (Q1, Q2)} affirmatively.

\begin{figure}
    \centering
    \includegraphics[width=0.46\textwidth]{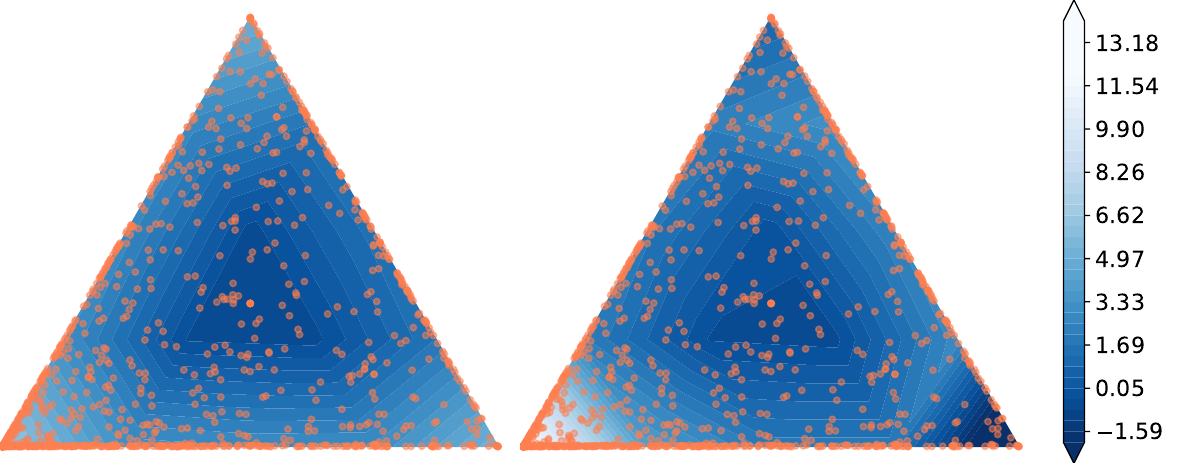}
    \caption{Simplex Distributions: Density of the topics spanning a 2-simplex from the NIPS dataset using {\bf (left)} Dirichlet and {\bf (right)} MSPN distributions. The more flexible MSPN distribution fits the topic distribution well and actually the lower-left topic better. (Best viewed in color).\label{fig:nips_simplex}}
    
\end{figure}

\textbf{Learning Simplex Distributions \textbf{(Q3)}:}
We considered data common in text and chemistry domains: proportional data, i.e., data lying on the probability simplex, the values are in $[0,1]$ and sum up to $1$. The Dirichlet distribution is arguably the most popular parametric distribution for this type of data. Hence, we used it as baseline. 

First, we considered the NIPS corpus, containing 1,500 documents over the 100 most frequent words. We ran Latent Dirichlet Allocation (LDA) \cite{Blei2012} with different number of topics (3,5,10,20,50) and used the topics found as the data for our experiments. Fig.~\ref{fig:nips_simplex} shows that the MSPN fits the density well and actually the lower-left topic  better.

\begin{figure*}[t]
\begin{minipage}{0.1\linewidth}
\vspace{-10pt}
    \includegraphics[height=1.4cm]{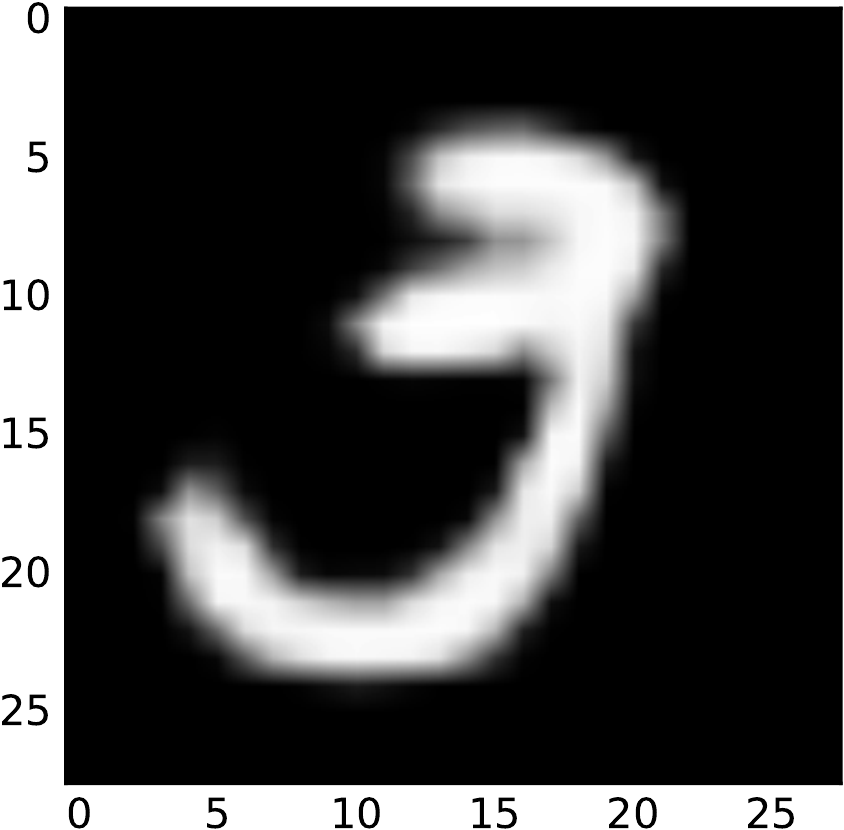}\\
    \includegraphics[height=1.4cm]{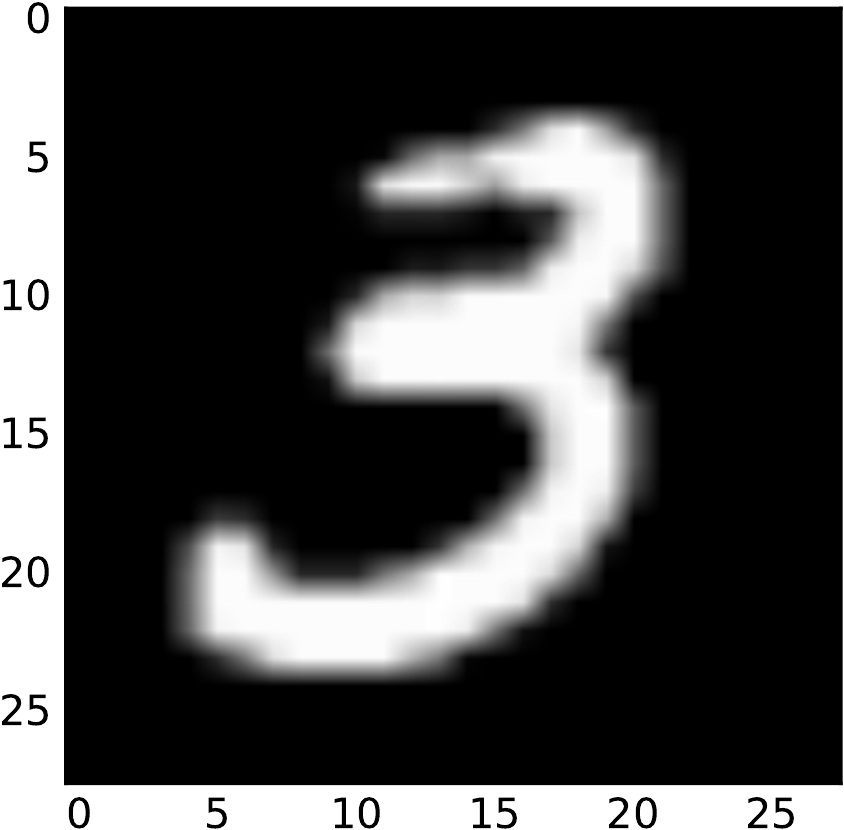}
\end{minipage}\hspace{0.1cm}\begin{minipage}{0.2\linewidth}
    \includegraphics[height=3cm]{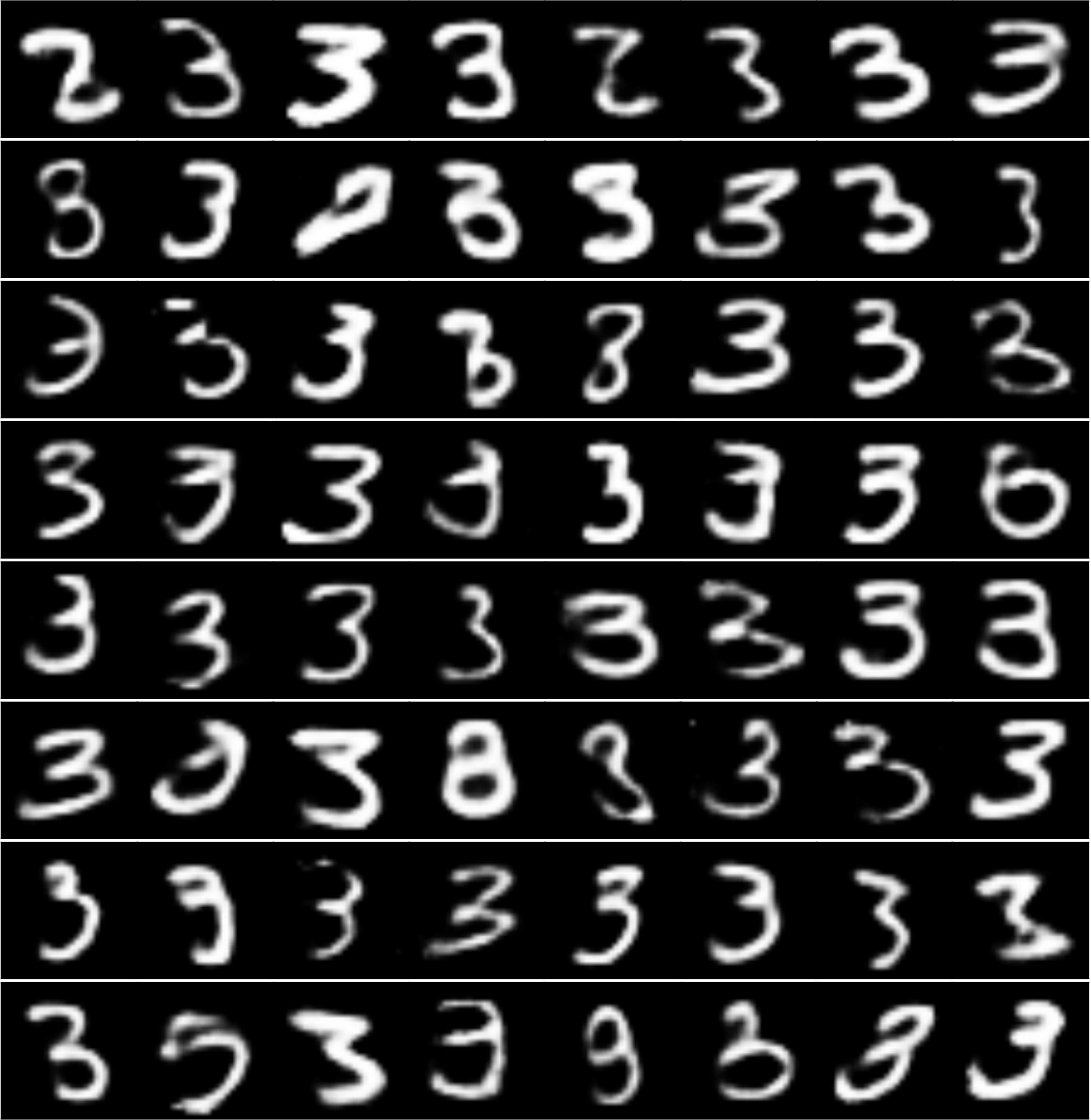}\\
\centering(a)
\end{minipage}\hspace{0.2cm}\begin{minipage}{0.2\linewidth}
    \includegraphics[height=3cm]{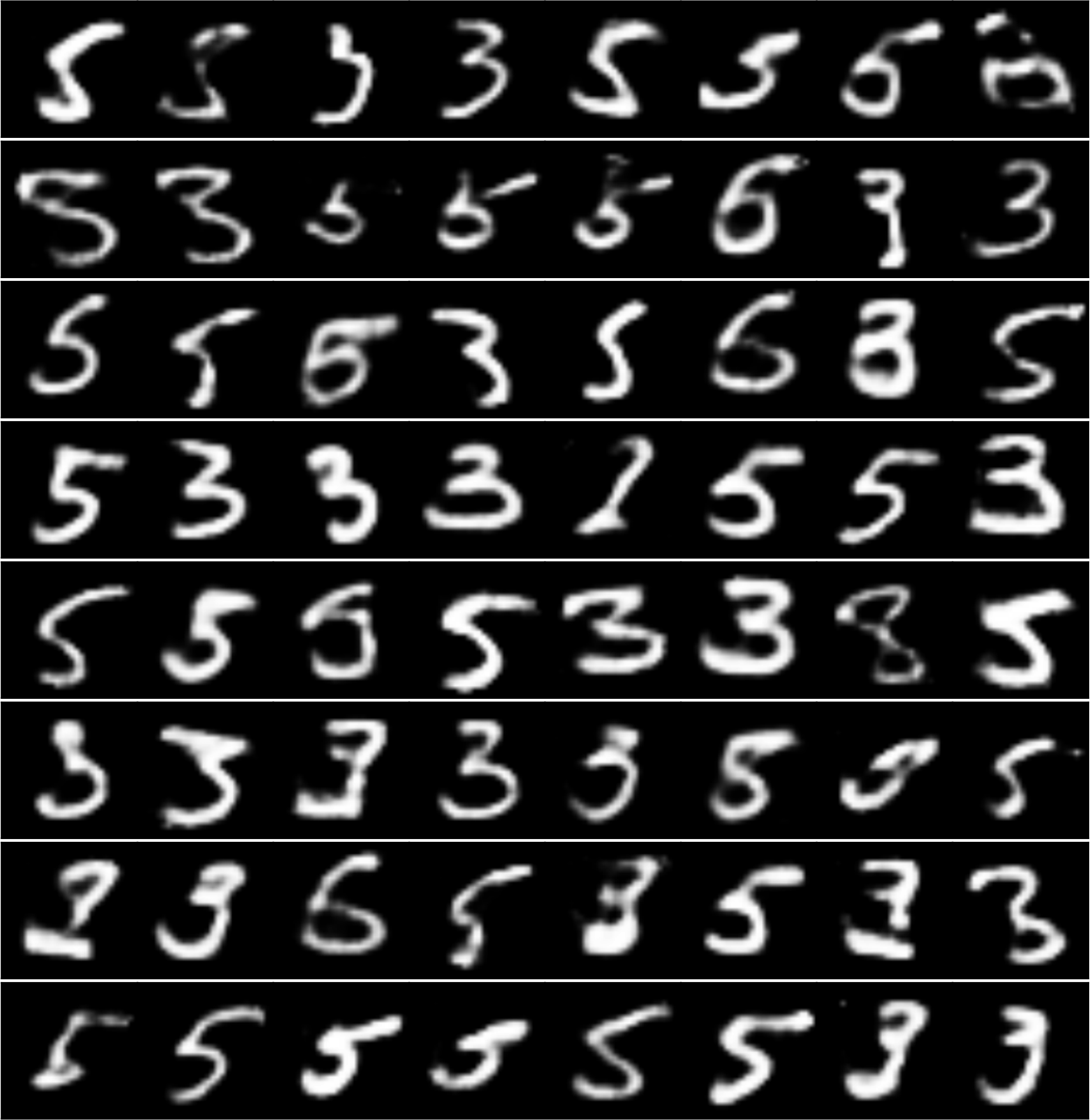}\\
\centering(b)
\end{minipage}\hspace{0.2cm}\begin{minipage}{0.2\linewidth}
    \includegraphics[height=3cm]{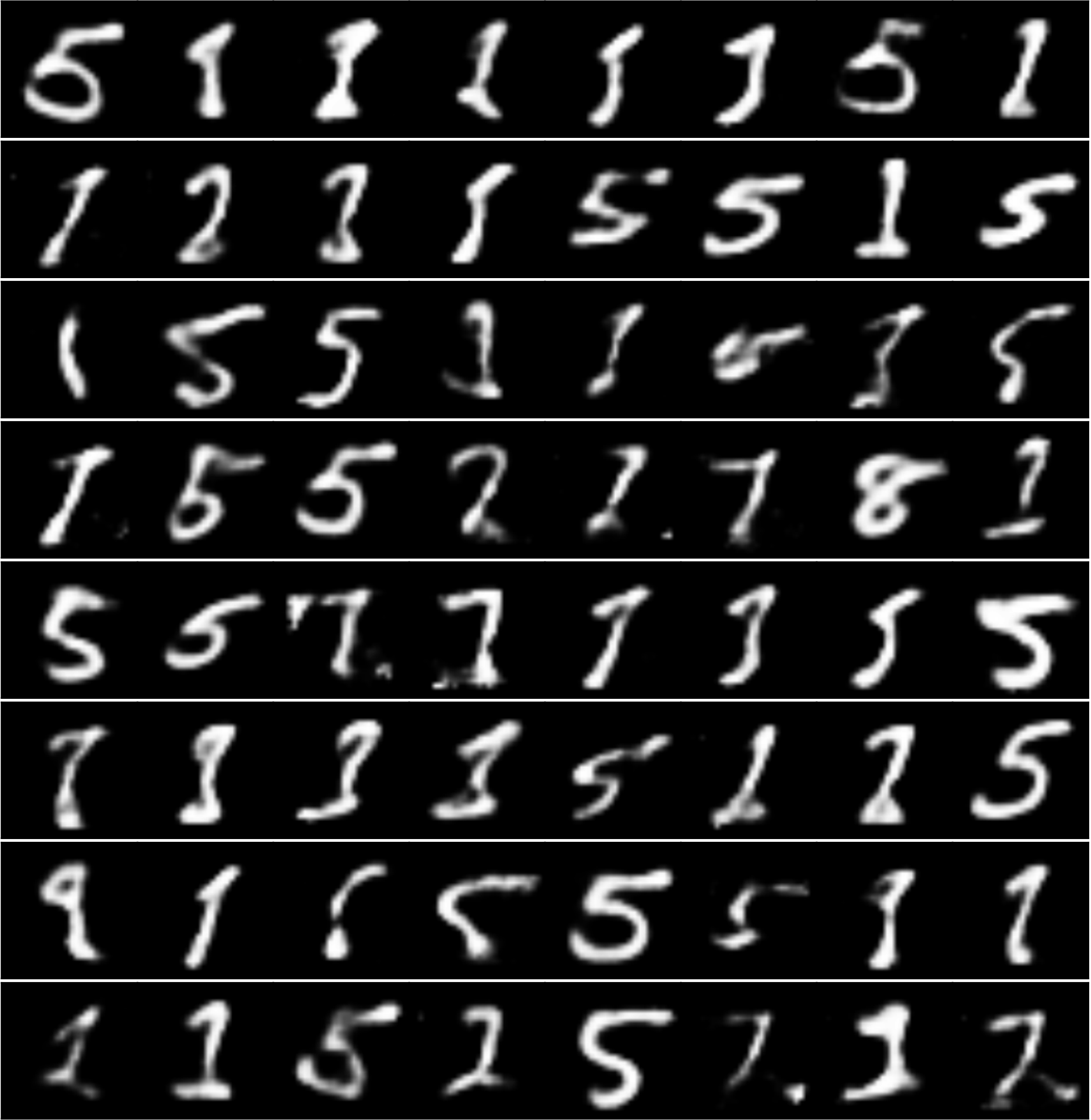}\\
    \centering(c)
\end{minipage}\hspace{0.2cm}\begin{minipage}{0.2\linewidth}
\vspace{0pt}
    \setlength{\tabcolsep}{-0.1pt}
    \begin{tabular}{rrrrr}
    \includegraphics[height=3cm]{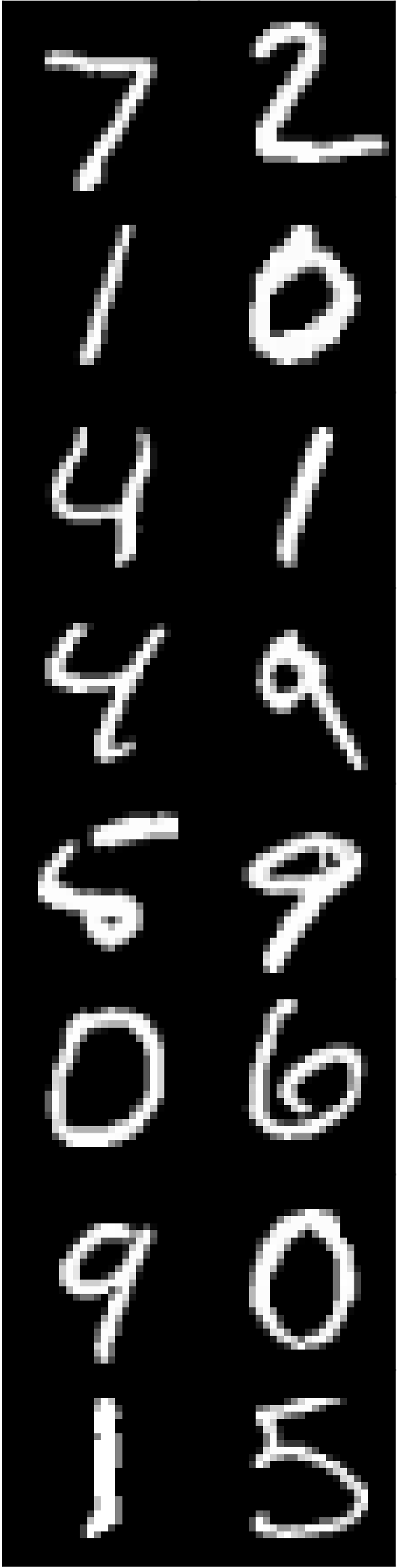}
    \includegraphics[height=3cm]{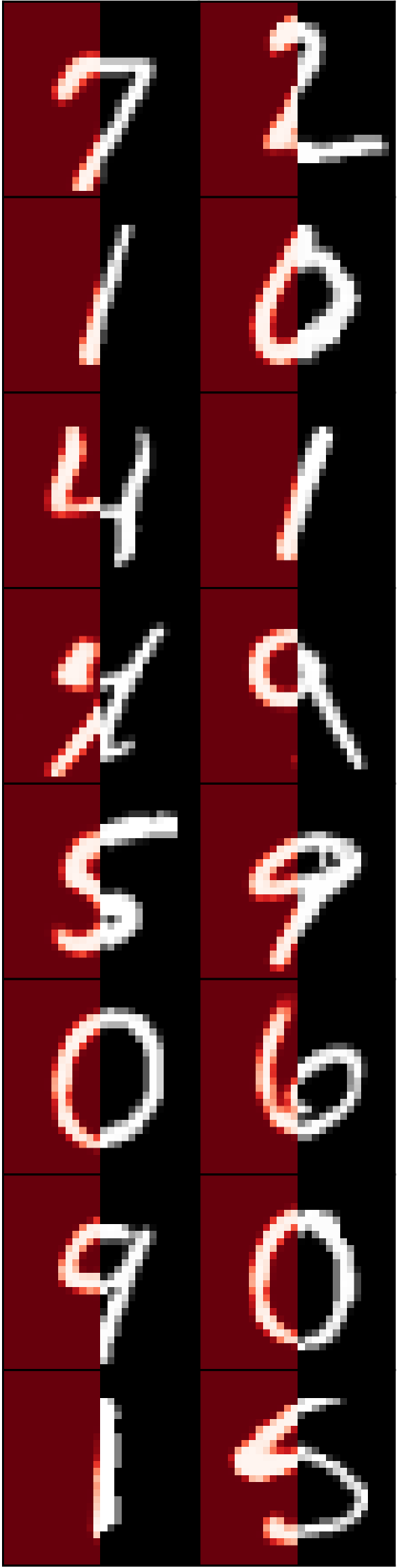}&
    \includegraphics[height=3cm]{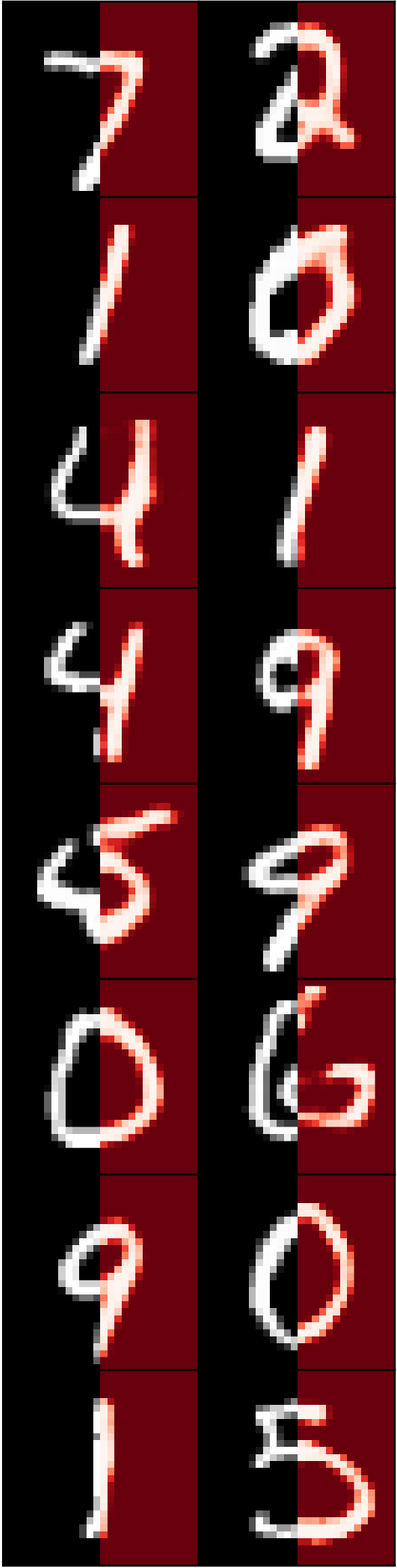}&
    \includegraphics[height=3cm]{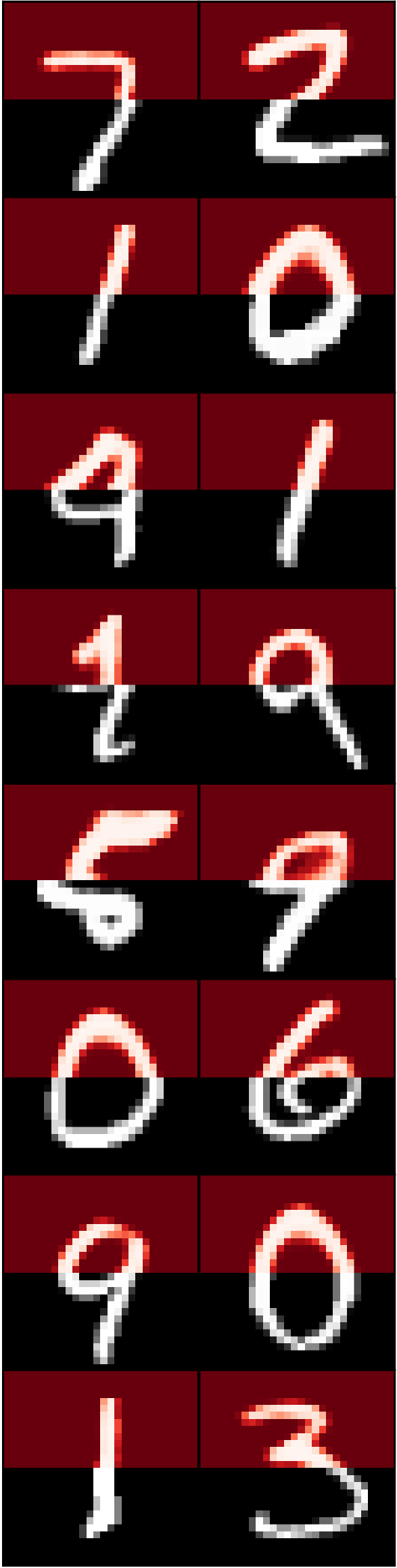}&
    \includegraphics[height=3cm]{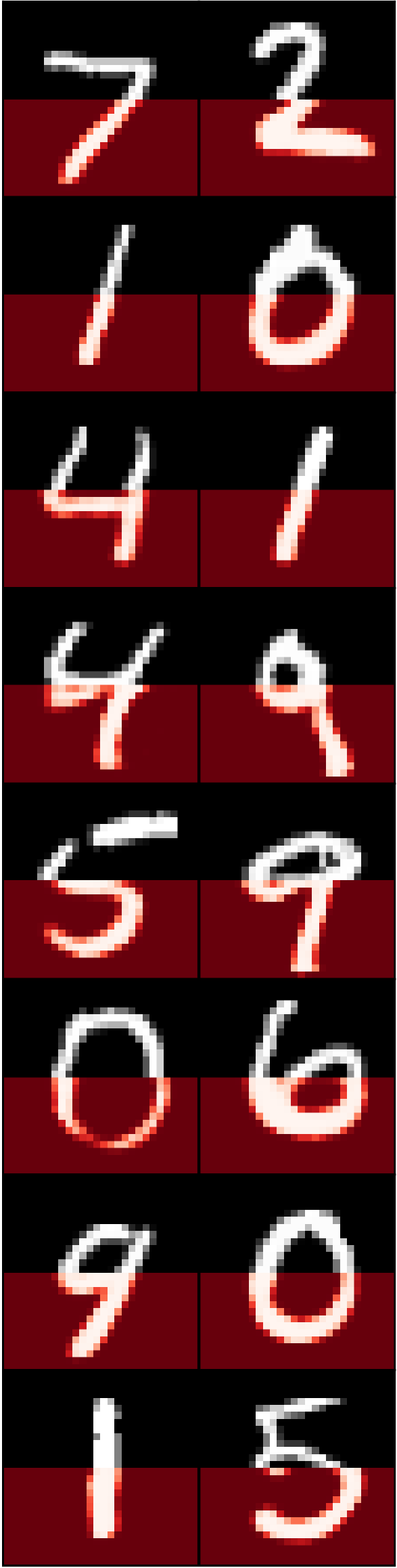}
    \end{tabular}\\

    \centering(d)
\end{minipage}
\caption[mnist]{Towards symbol grounding using MSPN.
    {\bf (a)} On the top left, the decoded sample predicted for the visual code  $\{0,0,1,0,1,1\}$ (corresponding to a ``3''), on the bottom left its training closest sample. 
    On the right, decoded conditional samples for the same visual code.
    {\bf (b,c)} Decoded conditional samples for codes in between classes ``3'' and ``5'' ($\{0,0,1,1,1,1\}$) as well as ``1'' and ``5'' ($\{1,0,1,1,1,0\}$), respectively.
    {\bf (d)} Some test images on the left and their MSPN reconstructions (left, right, up, down) on the right. The reconstructed parts are denoted by a red background. (Best viewed in color)}
    \label{fig:mnist}

\end{figure*}
\begin{table}[!t]
  \small
\centering
\setlength{\tabcolsep}{6pt}
\resizebox{\columnwidth}{!}{
\begin{tabular}{rr|rr}
\toprule
Dimension & \multicolumn{1}{r}{Dirichlet} & \multicolumn{1}{r}{MSPN(RDC,iso)} &
\multicolumn{1}{r}{MSPN(Grower,iso)} \\
\toprule
 &  \multicolumn{3}{l}{\textbf{NIPS + LDA}} \\
\toprule
3 & 2.045  ($\pm$ 0.297) & 4.071  ($\pm$ 0.66) & \textbf{4.333  ($\pm$ 0.627)} \\
5 & 7.311  ($\pm$ 0.406) & 10.376  ($\pm$ 0.671) & \textbf{10.419  ($\pm$ 0.711)} \\
10 & 25.047  ($\pm$ 0.787) & \textbf{35.927  ($\pm$ 1.755)} & 34.205  ($\pm$ 1.716) \\
20 & 69.668  ($\pm$ 2.014) & \textbf{109.222  ($\pm$ 4.179)} & 92.981  ($\pm$ 4.245) \\
50 & 245.008  ($\pm$ 3.573) & 338.477  ($\pm$ 6.976) & \textbf{349.259  ($\pm$ 9.916)} \\
 \toprule
 & \multicolumn{3}{l}{\textbf{Air Quality + Archetypes}} \\
\toprule
3 & 2.939 ($\pm$ 1.536) & 5.852  ($\pm$ 2.261) & \textbf{7.114 ($\pm$ 2.272)} \\
5 & 14.625  ($\pm$ 4.678) & \textbf{16.494  ($\pm$ 7.574)} & 15.099 ($\pm$ 4.888) \\
10 & 61.317  ($\pm$ 4.81) & 84.124  ($\pm$ 6.575) & \textbf{85.645 ($\pm$ 5.887)} \\
20 & 174.171  ($\pm$ 5.799) & 232.075  ($\pm$ 7.74) & \textbf{242.482 ($\pm$ 10.224)} \\
 \toprule
 & \multicolumn{3}{l}{\textbf{Hydrochemicals}} \\
\toprule
12 & 59.546  ($\pm$ 1.781) & 71.013  ($\pm$ 3.591) & \textbf{82.377  ($\pm$ 1.445)}\\ 
\toprule
{\bf wins over Dir.} & \multicolumn{1}{c}{\bf -} & \multicolumn{1}{c}{\bf 10/10} & \multicolumn{1}{c}{\bf 10/10} \\
{\bf wins} & \multicolumn{1}{c}{\bf 0/10} & \multicolumn{2}{c}{\bf 10/10}
\end{tabular}
}
\caption[datasets]{Average test set log likelihoods (the higher, the better) on proportional data; best results bold. Clearly, MSPNs  outperform the less flexible Dirichlet distribution, even without information about the statistical type (RDC, iso). 
\label{tab:simplex}}
\end{table}

Then, we investigated the Air Quality dataset\footnote{\url{https://archive.ics.uci.edu/ml/datasets/Air+Quality}. We used only complete instances and ignored the time features as well as feature C6H6 that has many missing instances} containing 6,941 measurements for 12 features about air composition. We ran Archetypal Analysis \cite{Cutler1994,Thurau2012} for 3, 5, and 10 archetypes and extracted the archetypical convex reconstructions of the original data.

We also considered the hydro-chemical dataset of \citeauthor{Tolosana-Delgado2005}~(\citeyear{Tolosana-Delgado2005}), containing 485 observations of 14 measurements of different chemicals for the Llobregat river in Spain. The relative concentrations are used to fit MSPNs and the Dirichlet distributions. The 10-fold cross-validated mean log-likelihoods for all models on the three datasets are summarized in Table~\ref{tab:simplex}.
As one can see, in all cases MSPNs can capture the distribution on the simplex better than the Dirichlet. This is to be expected as MSPNs can capture more complex (in)dependencies, whereas the Dirichlet makes stronger independence assumptions. All simplex experiments together answer {\bf (Q3)} affirmatively.

\textbf{Leveraging symbolic-semantic information~\textbf{(Q4)}:}
Symbol grounding---the problem of how symbols get their meanings---is at the heart of AI, and we explored MSPNs as a step towards tackling this classicial AI problem.
More precisely, we considered the 28$\times$28 MNIST gray digit images. We represented the digit as 16 continuous features extracted from an autoencoder (AE) trained on the MNIST training split: we stacked two layers of 256 and 128 rectifier neurons for both the encoder and the decoder and trained them for 200 epochs using \textsf{adam} as optimizer (learning rate $0.002$, $\beta_{1}$ and $\beta_{2}$ coefficients set to $0.9$ resp.~$0.999$,and no learning rate decay). To create a hybrid dataset, we then augmented MNIST 
with symbolic semantic information encoded as binary codes. 
Each bit of the code is 1 if a digit contains one of the following visual features:
(i) a vertical stroke (true for 1, 4 and 7),
(ii) a circle (0, 6, 8 and 9),
(iii) a left curvy stroke (2, 3, 5, 8 and 9),
(iv) a right curvy stroke (5 and 6),
(v) a horizontal stroke (7, 2, 3, 4, and 5),
(vi) a double curve stroke (3 and 8).
That is, each class is encoded by a 6-bit code. For instance, images representing a ``3'' are assigned the code $(0,0,1,0,1,1)$ while  $(0,0,1,1,1,0)$ corresponds to ``5''. Additionally, we considered the original class variable $C$ as a third piece of information. Let $\X$ denote the continuous embedding variables, $\Y$ the additional 6 binary symbolic features, and $C$ the categorical class variable.

\begin{figure}
    \centering
    \includegraphics[width=0.48\columnwidth]{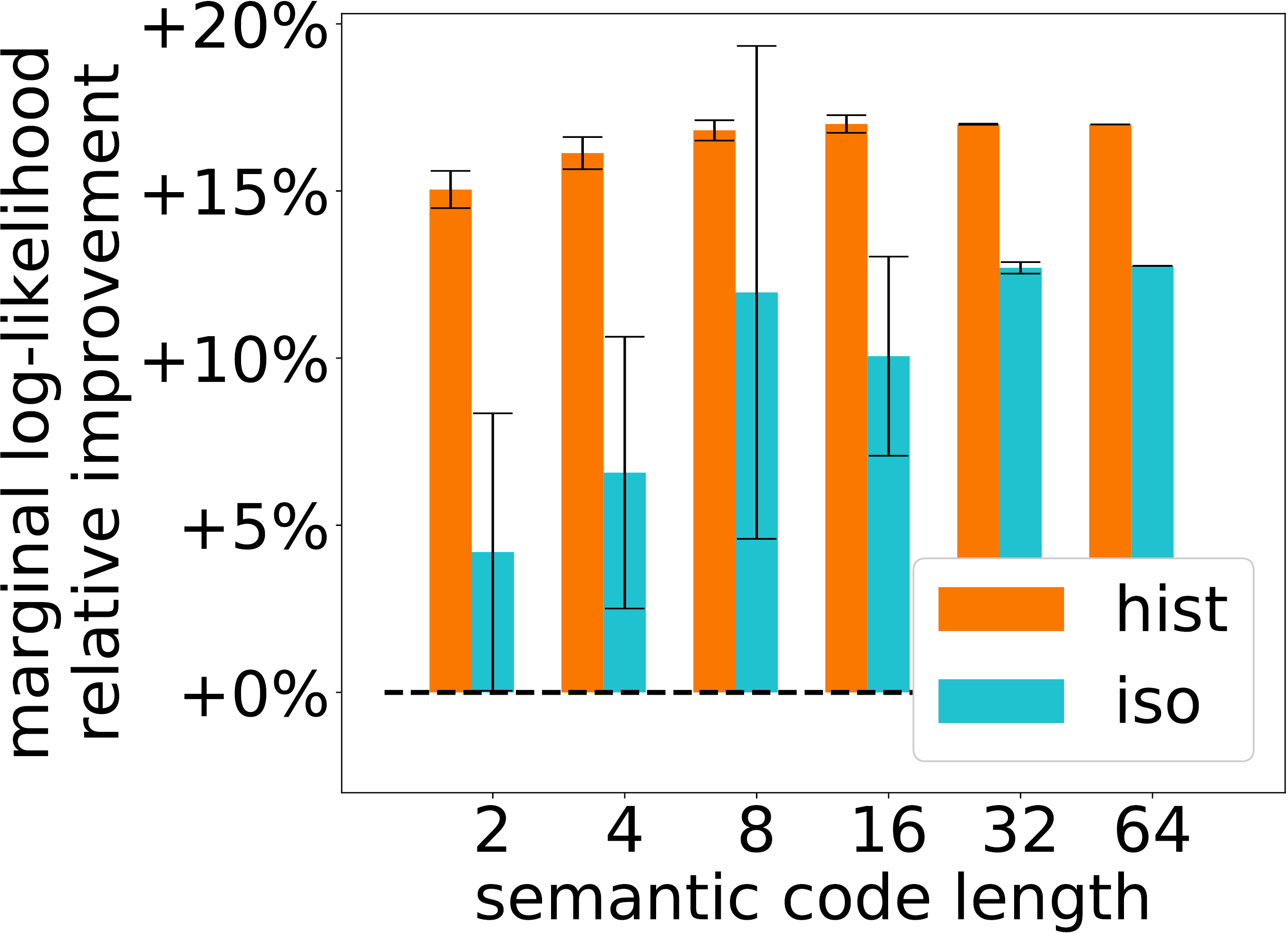}
    \includegraphics[width=0.48\columnwidth]{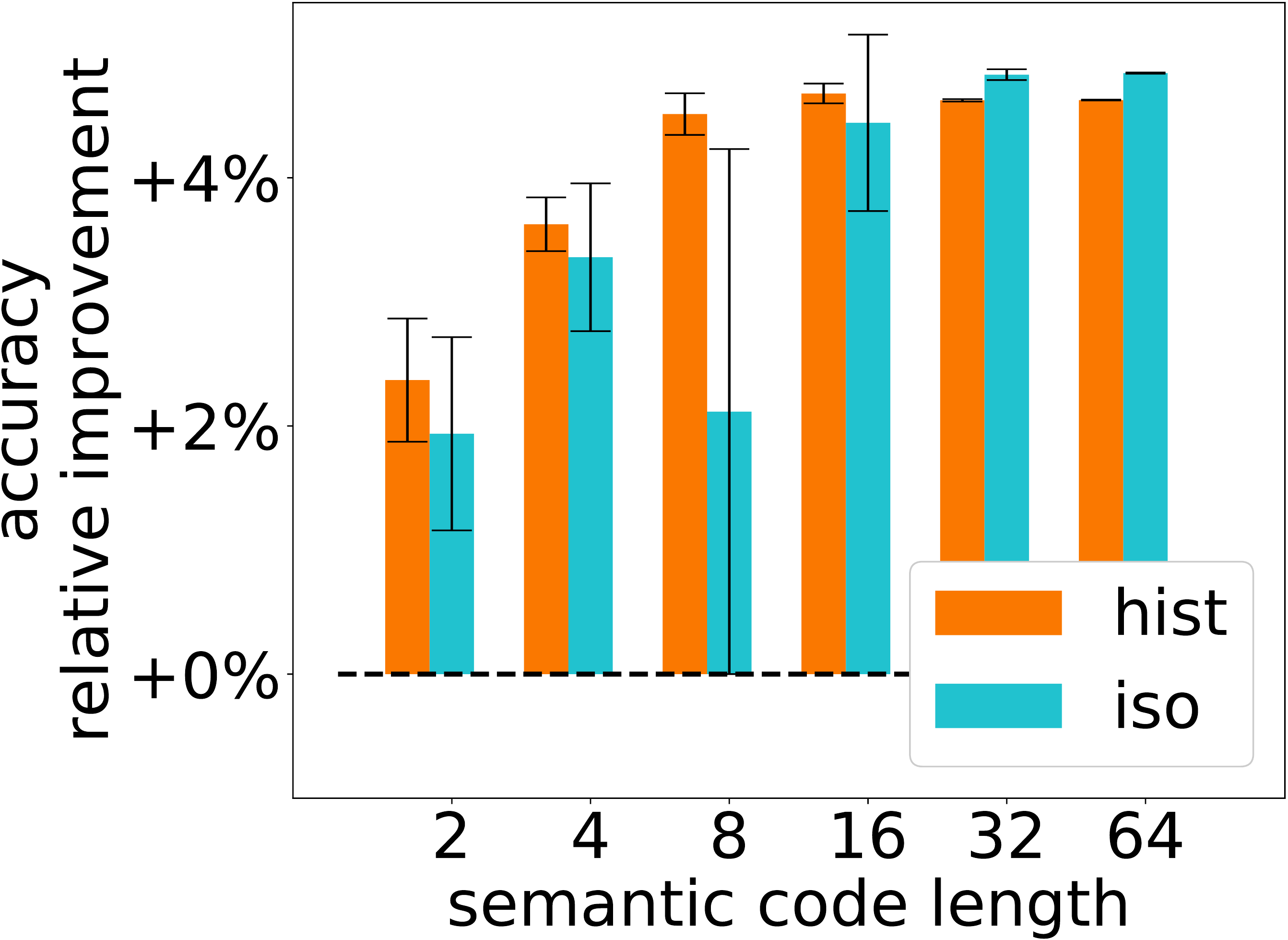}
    \caption{Average relative improvement over ten trials (y axis) for {\bf (left)} the marginal test log-likelihood $P(\X)$ and {\bf (right)} for the class accuracy based on $P(C)$ of MSPNs learned an autoencodings augmented with semantic class codes of increasing length (x axis). (Best viewed in color)}
    \label{fig:mnist-bars-dens}
\end{figure}

In a first experiment, we trained an MSPN on a 10000 subsample of the augmented MNIST training data to model $P(\X, \Y)$,  setting $\eta = 200$ and $\Delta = 1$, $k=20$.
Then, we evaluated on the augmented MNIST test split whether the learned MSPN had captured the non explicit dependencies between the three different feature domains.
First, we predict $\x^{*}=\argmax_{\x} P(\x|\Y=\y_{c})$, for each visual code $\y_{c}$ belonging to class $c\in C$.
Fig.~\ref{fig:mnist} (a) visualizes the prediction $\x^{*}$ as decoded by the autoencoder back in pixel space.
As one can see, the MSPN is not only able to recover the correct class but also does not simply memorize a training sample.
An additional visual proof is provided by conditional sampling: after propagating bottom-up the evidence for an observed code $\y_{c}$, we sample a configuration $\x$ (applying \citeauthor{Vergari2016a}'s~(\citeyear{Vergari2016a}) top down approach).
Decoded samples clearly belong to the class $c$, cf. Fig.~\ref{fig:mnist} (a).
Then, to evaluate how good the MSPN was able to glue the continuous and binary domains, we performed conditional sampling starting from unseen visual codes.
For instance, for the code $(0, 0, 1,1,1,1)$, we expect a digit in between a ``3'' and a ``5'', since it is merging the visual codes of these two classes.
Fig.~\ref{fig:mnist}~(b) confirms this: decoded samples belong to either class or are closely ``in between'' them.
Similarly, Fig.~\ref{fig:mnist}~(c) shows samples conditioned on code $\{1,0,1,1,1,0\}$, in between class 5 and 1.

Next, we investigated how much symbol groundings can be helpful for density estimation and classification.
On the MNIST test split, we investigated the benefit of using visual codes $\Y$ of length 2,4,8,16,32,64. 
We measured the improvement of the marginal likelihood $P(\X)$ resp.~the classification accuracy based on $P(C)$ of an MSPN $\mathcal{B}$ trained on $(\X, \Y, C)$ over an MSPN $\mathcal{A}$ trained only over $(\X, C)$: $({\ell_{\mathcal{M}} - \ell_{\mathcal{B}}})\slash{\ell_{\mathcal{B}}}\cdot 100$ for both measures $\ell$. The results are summarized in Fig.~\ref{fig:mnist-bars-dens}.
As one can see, increasing the number of symbolic features positively improves both the marginal likelihood over $\X$ and the classification performance. Note that for computing $P(\X)$ and to predict $c^{*}=\argmax_{c}P(c|\X)$, one has to marginalize over $\Y$, which cannot be done efficiently using classical mixed graphical models.

Finally, we employed MSPNs for MNIST reconstruction.
We processed the original images as two halves---left (l) and right (r), up (u) and down (d)---and encoded each half into 16 continuous features by learning one autoencoder independently for each one of them. Note that each variable set $X_{l}$, $X_{r}$, $X_{u}$ and $X_{d}$ forms a domain with a different distribution.
Then, we learned MSPNs for $P(\X_{l}, \X_{r})$ and $P(\X_{u}, \X_{d})$. We performed MPE inference to predict one half of a test image given the complementary one, e.g. left from right.
Predicted samples are shown in Fig.~\ref{fig:mnist} (d). As one can see, the reconstructions are indeed very plausible.
This suggests that MSPNs are a valuable tool to effectively learn distributions and make predictions across different domains. 
All the experiments on leveraging symbolic-semantic information together answer {\bf (Q4)} affirmatively.

\begin{figure}[t]
    \centering
    \includegraphics[width=0.6\columnwidth]{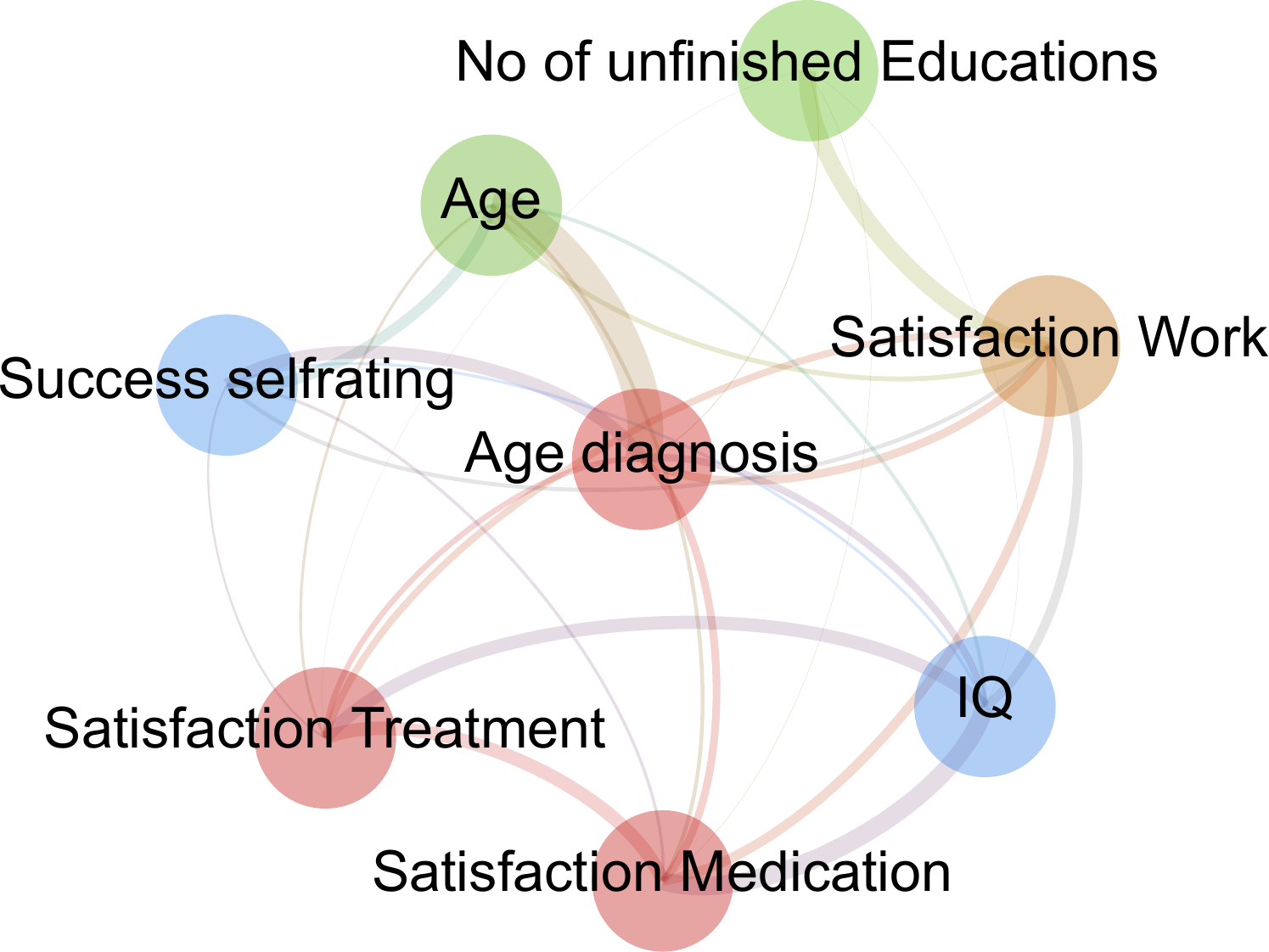}

    \caption{Visualizing the gist of the Autism MSPN using normalized mutual information (the thicker, the higher). There is a strong relationship between {\it Age} and {\it Age of diagnosis} (one cannot be diagnosed before being born) as well as between {\it Satisfaction Work} and {\it No of unfinished Education} (the more unfinished educations the less satisfied at work). This is akin to the results of~\cite{Haslbeck2015}, who estimated a pairwise mixed graphical model with known parametric forms from the exponential family~\cite{Yang2014} using regularized neighborhood regression. Node colors encode different feature groups: Demographics (green), Psychological (blue), Social Environment (organge) and Medical (red). (Best viewed in color)   \label{fig:mutualinfo}
}
    
\end{figure}

\textbf{Mixed Mutual Information~\textbf{(Q5)}:}
Recall, an MSPN encodes a polynomial over leaf piecewise polynomials. 
Consequently, one can employ a symbolic solver to evaluate the overall network polynomial to easily compute information-theoretic measures that would be difficult to compute otherwise, in particular for hybrid domains. To illustrate this for MSPNs, we consider computing mutual information (MI) in hybrid domains. MI also provides a way to extract the gist of MSPNs as it highlights relevant variable associations only. Fig.~\ref{fig:mutualinfo} shows the MI network induced over the Autism Dataset~\cite{Deserno2016}, which reflects natural semantic connections. 
This not only answers \textbf{(Q5)} affirmatively but also indicates that MSPNs may pave the way to automated mixed statisticians: the MI together with the tree structure of MSPNs can automatically be compiled into textual descriptions of the model; and interesting avenue for future work. 
To summarize our experimental results as a whole, all questions {\bf (Q1)}-{\bf (Q5)} can be answered affirmatively. 

\section{Conclusions}
We introduced Mixed Sum-Product Networks (MSPNs), a novel combination of nonparametric probability distributions and deep probabilistic models.
In contrast to classical shallow mixed graphical models, they provide effective learning, a range of tractable inferences and enhanced interpretability.
Our experiments demonstrate that MSPNs are competitive to parameterized distributions as well as mixed graphical models and make previously difficult---if not impossible--- to compute queries easy. 
Hence, they allow users to train multivariate mixed distributions more easily than previous approaches across a wide range of domains.

MSPNs suggest several avenues for future work: from learning boosted and mixtures of MSPNs along with exploring other nonparametric leaves such KDE, other mixed graphical models, and variational autoencoders, extending them to other instances of arithmetic circuits~\cite{choiD17}, and making use of weighted model integration solvers for capturing more complex types of queries~\cite{Belle2015a,morettinPS17}.
Probably the most interesting avenue is to turn MSPNs into automated statisticians, able to predict the statistical type of a variable---is it continuous or ordinal?---and ultimately its parametric form---is it Gaussian or Poisson~\cite{Valera2017a}?

{\bf Acknowledgements:} This work is motivated and partly supported by the BMEL/BLE project DePhenSe, FKZ 313-06.01-28-1-82.047-15. AM has been supported by Deutsche Forschungsgemeinschaft (DFG) within the Collaborative Research Center SFB 876 "Providing Information by Resource-Constrained Analysis", projects B4.  SN has been supported by the CwC Program Contract W911NF-15-1-0461 with the US Defense Advanced Research Projects Agency (DARPA) and the Army Research Office (ARO).

\bibliography{referomnia}
\end{document}